\documentclass[10pt,twocolumn,letterpaper]{article}

\usepackage{iccv}
\usepackage{times}
\usepackage{epsfig}
\usepackage{graphicx}
\usepackage{amsmath}
\usepackage{amssymb}
\usepackage{footnote}
\usepackage{subfig}

\makeatletter
\providecommand\phantomcaption{\caption@refstepcounter\@captype}
\makeatother

\usepackage[pdfstartview=FitH,pagebackref=true,breaklinks=true,colorlinks,bookmarks=false,linkcolor=blue,citecolor=blue]{hyperref}

\iccvfinalcopy 

\def\iccvPaperID{1554} 

\ificcvfinal\fi
\begin{document}

\title{Unsupervised Holistic Image Generation from Key Local Patches}

\author{Donghoon Lee$^1$, Sangdoo Yun$^1$, Sungjoon Choi$^1$, Hwiyeon Yoo$^1$, Ming-Hsuan Yang$^2$, and Songhwai Oh$^1$\\
$^1$Electrical and Computer Engineering and ASRI, Seoul National University, Korea\\
$^2$Electrical Engineering and Computer Science, University of California at Merced\\
{\tt\small donghoon.lee@cpslab.snu.ac.kr, yunsd101@snu.ac.kr, sungjoon.choi@cpslab.snu.ac.kr} \\
{\tt\small hwiyeon.yoo@cpslab.snu.ac.kr, mhyang@ucmerced.edu, songhwai@snu.ac.kr}}


\maketitle

\begin{abstract}
We introduce a new problem of generating an image based on a small number of key local patches without any geometric prior.
In this work, key local patches are defined as informative regions of the target object or scene.
This is a challenging problem since it requires generating
realistic images and predicting locations of parts at the same time.
We construct adversarial networks to tackle this problem.
A generator network generates a fake image as well as a mask based on the encoder-decoder framework.
On the other hand, a discriminator network aims to detect fake images.
The network is trained with three losses to consider spatial, appearance, and adversarial information.
The spatial loss determines whether the locations of predicted parts are correct.
Input patches are restored in the output image without much modification due to the appearance loss.
The adversarial loss ensures output images are realistic.
The proposed network is trained without supervisory signals since no labels of key parts are required.
Experimental results on six datasets demonstrate that the proposed algorithm
performs favorably on challenging objects and scenes.
\end{abstract}

\vspace{-2mm}
\section{Introduction}
\vspace{-2mm}
The goal of image generation is to construct images that are as barely distinguishable
from target images which may contain general objects, diverse scenes, or human drawings.
Synthesized images can contribute to a number of applications such as the
image to image translation \cite{isola2016image},
image super-resolution \cite{ledig2016photo},
3D object modeling \cite{wu2016learning},
unsupervised domain adaptation \cite{liu2016coupled},
domain transfer \cite{yoo2016pixel},
future frame prediction \cite{vondrick2016generating},
image inpainting \cite{yeh2016semantic},
image editing \cite{zhu2016generative}, and
feature recovering of astrophysical images \cite{schawinski2017generative}.
%

\begin{figure}[t]
    \centering
    \includegraphics[width=1.0\linewidth]{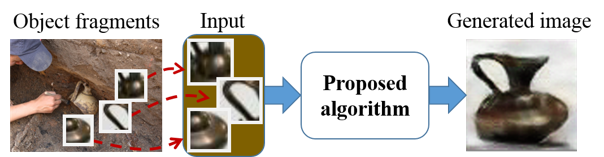}
    \caption{
The proposed algorithm is able to
synthesize an image from key local patches without geometric priors, e.g.,
restoring broken pieces of ancient ceramics found in ruins.
Convolutional neural networks are trained to predict locations of input patches
and generate the entire image based on adversarial learning.
}
\label{fig:teaser}
\end{figure}

In this paper, we introduce a new image generation problem, in which a
whole image is generated conditioned on parts of an image.
The objective of this work, as shown in Figure \ref{fig:teaser},
is to generate an image based on a small number of local patches without geometric priors.
This problem is more complicated than conventional image generation tasks as it
entails to achieve three goals simultaneously.

First, spatial arrangements of input patches need to
be inferred since the data does not contain explicit information about the location.
To tackle this issue, we assume that inputs are key local patches which are
informative regions of the target image.
Therefore, the algorithm should learn the spatial relationship between key parts
of an object or scene.
Our approach obtains key regions without any supervision
such that the whole algorithm is developed within the unsupervised learning framework.

Second, we aim to generate an image while preserving the key local patches.
As shown in Figure \ref{fig:teaser}, the appearances input patches are included in the generated image without significant modification.
In other words, the inputs are not directly copied to the output image.
It allows us to create images more flexibly such that we can combine key patches of
different objects as inputs.
In such cases, input patches must be deformed by considering each other.

Third and most importantly, the generated image should look closely to
a real image in the target category.
Unlike the image inpainting problem,
which mainly replaces small regions or eliminates minor defects,
our goal is to reconstruct a holistic image based on limited appearance information
contained in a few patches.

To address the above issues, we adopt the adversarial learning scheme \cite{goodfellow2014generative} in this work.
The generative adversarial network (GAN) contains two networks which are trained based on the min-max game of two players.
A generator network typically generates fake images and aims to fool a discriminator,
while a discriminator network seeks to distinguish fake images from real images.
In our case, the generator network is also responsible for predicting the
locations of input patches.
Based on the generated image and predicted mask,
we design three losses to train the network:
a spatial loss, an appearance loss, and an adversarial loss,
corresponding to the aforementioned issues, respectively.

While a conventional GAN is trained in an unsupervised manner,
some recent methods formulate it in a supervised manner by using labeled information.
For example, a GAN is trained with a dataset that have 15 or more joint positions of birds \cite{reed2016learning}.
Such labeling task is labor intensive since GAN-based algorithms need a large amount
of training data to achieve high-quality results.
In contrast, experiments on six challenging datasets that contain different objects and scenes,
such as faces, cars, flowers, ceramics, and waterfalls,
demonstrate that the proposed unsupervised algorithm can
generate realistic images and predict part locations well.
In addition, even if inputs contain parts from different objects,
our algorithm is able to generate reasonable images.

The main contributions are as follows.
%
%
%
%
%
%
First, we introduce a new problem to render realistic image conditioned on the appearance information of a few key patches.
Second, we develop a generative network to jointly predict the mask and the image without supervision to address the defined problem.
Third, we propose a novel objective function using additional fake images to strengthen the discriminator network.
Finally, we provide new datasets that contain challenging objects and scenes.

\section{Related Work}
\vspace{-2mm}
{\flushleft{\bf Image Generation.}}
Image generation is an important problem that has been studied extensively in computer vision.
With the recent advances in deep convolutional neural networks~\cite{krizhevsky2012imagenet,simonyan2014very},
numerous image generation methods have achieved the state-of-the-art results.
Dosovitskiy et al.~\cite{dosovitskiy2015learning} generate 3D objects by learning transposed convolutional neural networks.
In~\cite{kingma2013auto}, Kingma et al. propose a method based on
the variational inference for the stochastic image generation.
An attention model is developed by Gregor et al.~\cite{gregor2015draw}
to generate an image using a recurrent neural network.
Recently, the stochastic PixelCNN~\cite{van2016conditional} and PixelRNN~\cite{oord2016pixel} are introduced to generate images sequentially.

The generative adversarial network~\cite{goodfellow2014generative} is proposed
for generating sharp and realistic images based on two competing networks: a generator and
a discriminator.
Numerous methods~\cite{salimans2016improved,zhao2016energy} have been proposed to improve the stability of the GAN.
Radford et al.~\cite{radford2015unsupervised} propose
deep convolutional generative adversarial networks (DCGAN)
with a set of constraints to generate realistic images effectively.
Based on the DCGAN architecture, Wang et al.~\cite{wang2016generative}
develop a model to generate the style and structure of indoor scenes (SSGAN),
and Liu et al.~\cite{liu2016coupled} present a coupled GAN which learns a joint distribution
of multi-domain images, such as color and depth images.

{\flushleft{\bf Conditional GAN.}}
Conditional GAN approaches~\cite{mirza2014conditional,reed2016generative,zhang2016stackgan} are developed to control the image generation process with label information.
Mizra et al.~\cite{mirza2014conditional} propose a class-conditional GAN which uses discrete class labels as the conditional information.
The GAN-CLS~\cite{reed2016generative} and StackGAN~\cite{zhang2016stackgan} embed a text describing an image into the conditional GAN to generate an image corresponding to the condition.
On the other hand, the GAWWN~\cite{reed2016learning} creates numerous plausible images based on the location of key points or an object bounding box.
In these methods, the conditional information, e.g., text, key points, and bounding boxes, is provided in the training data.
However, it is labor intensive to label such information since deep generative models require a large amount of training data.
In contrast, key patches used in the proposed algorithm are obtained without the necessity of human annotation.

Numerous image conditional models based on GANs have been introduced recently~\cite{ledig2016photo, zhu2016generative, yoo2016pixel, yeh2016semantic, pathak2016context, li2016precomputed, shrivastava2016learning,isola2016image}.
These methods learn a mapping from the source image to target domain, such as image super-resolution~\cite{ledig2016photo}, user interactive image manipulation~\cite{zhu2016generative}, product image generation from a given image~\cite{yoo2016pixel}, image inpainting~\cite{yeh2016semantic,pathak2016context}, style transfer~\cite{li2016precomputed} and realistic image generation from synthetic image \cite{shrivastava2016learning}.
Isola et al.~\cite{isola2016image} tackle the image-to-image translation problem including various image conversion examples
such as day image to night image, gray image to color image, and sketch image to real image, by utilizing the U-net~\cite{ronneberger2015u} and GAN.
In contrast, the problem addressed in this paper is the holistic image generation based on only a small number of local patches.
This challenging problem cannot be addressed by existing image conditional methods as the domain of the source and target images are different.

\begin{figure*}[t]
    \centering
    \includegraphics[width=1.0\linewidth]{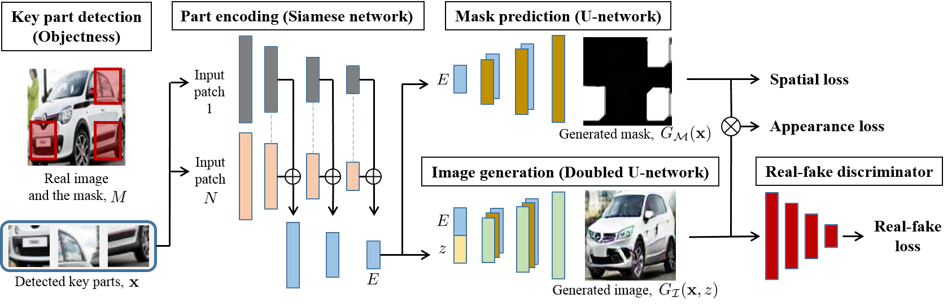}
    \caption{
    Proposed network architecture.
    A bar represents a layer in the network.
    Layers of the same size and the same color have the same convolutional feature maps.
    Dashed lines in the part encoding network represent shared weights.
    In addition, $E$ denotes an embedded vector and $z$ is a random noise vector.
}
\label{fig:proposedNet}
\end{figure*}

{\flushleft{\bf Unsupervised Image Context Learning.}}
Unsupervised learning of the spatial context in an image~\cite{doersch2015unsupervised,noroozi2016unsupervised,pathak2016context}
has attracted attention to learn rich feature representations without human annotations.
Doersch et al.~\cite{doersch2015unsupervised} train convolutional neural networks to predict the relative position between two neighboring patches in an image.
%
The neighboring patches are selected from a grid pattern based on the image context.
To reduce the ambiguity of the grid in \cite{doersch2015unsupervised}, Noroozi et al.~\cite{noroozi2016unsupervised} divide the image into a large number of tiles,
shuffle the tiles, and then learn a convolutional neural network to solve the jigsaw puzzle problem.
Pathak et al.~\cite{pathak2016context} address the image inpainting problem which predicts missing pixels in an image, by training a context encoder.
Through the spatial context learning, the trained networks are successfully applied to various applications such as object detection, classification and semantic segmentation.
However, discriminative models \cite{doersch2015unsupervised, noroozi2016unsupervised} can only infer the spatial arrangement of image patches,
and the image inpainting method \cite{pathak2016context} requires the spatial information of the missing pixels.
In contrast, we propose a generative model which is capable of not only inferring the spatial arrangement of input patches but also generating the entire image.

\vspace{-2mm}
\section{Proposed Algorithm}
\vspace{-2mm}
Figure \ref{fig:proposedNet} shows the structure of the proposed network for image generation from a few patches.
It is developed based on the concept of adversarial learning, where a generator and a discriminator compete with each other \cite{goodfellow2014generative}.
However, in the proposed network, the generator has two outputs: the predicted mask and generated image.
Let $G_\mathcal{M}$ be a mapping from $N$ observed images $\mathbf{x}=\{x_1,...,x_N\}$ to a mask $M$, $G_\mathcal{M} : \mathbf{x}\rightarrow M$.\footnote{
Here, $\mathbf{x}$ is a set of image patches resized to the same width and height suitable for the proposed network and $N$ is the number of image patches in $\mathbf{x}$.}
Also let $G_\mathcal{I}$ be a mapping from $\mathbf{x}$ and a random noise vector $z$ to an output image $y$, $G_\mathcal{I} : \{\mathbf{x},z\}\rightarrow y$.
These mappings are performed based on three networks: a part encoding network, a mask prediction network, and an image generation network.
The discriminator $D$ is based on a convolutional neural network which aims to distinguish the real image from the image generated by $G_\mathcal{I}$.

We use three losses to train the network.
The first loss is the spatial loss $\mathcal{L}_S$.
It compares the inferred mask and real mask which represents the cropped region of the input patches.
The second loss is the appearance loss $\mathcal{L}_A$, which maintains input key patches in the generated image without much modification.
The third loss is the adversarial loss $\mathcal{L}_R$ to distinguish fake and real images.
%
The whole network is trained by the following min-max game:
\begin{equation}\label{eq:loss}
  \min_{G_\mathcal{M},G_\mathcal{I}}\max_D \mathcal{L}_{R}(G_\mathcal{I},D)
                                         + \lambda_1 \mathcal{L}_{S}(G_\mathcal{M})
                                         + \lambda_2 \mathcal{L}_{A}(G_\mathcal{M},G_\mathcal{I}),
\end{equation}
where $\lambda_1$ and $\lambda_2$ are weights for the spatial loss and the appearance loss, respectively.
\vspace{-2mm}
\subsection{Key Part Detection}
\vspace{-2mm}
\begin{figure}[t]
    \centering
    \includegraphics[width=1.0\linewidth]{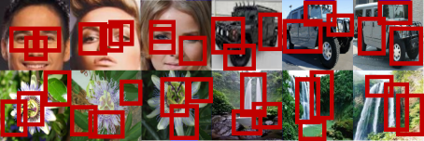}
    \caption{
    Examples of detected key patches on faces \cite{liu2015faceattributes},
    vehicles \cite{krause20133d}, flowers \cite{nilsback2008automated}, and waterfall scenes.
    Three regions with top scores from the EdgeBox algorithm are shown in red boxes after pruning candidates of an extreme size or aspect ratio.
}
\label{fig:keyPatches}
\end{figure}
\begin{figure*}[!t]
    \centering
    \includegraphics[width=0.9\linewidth]{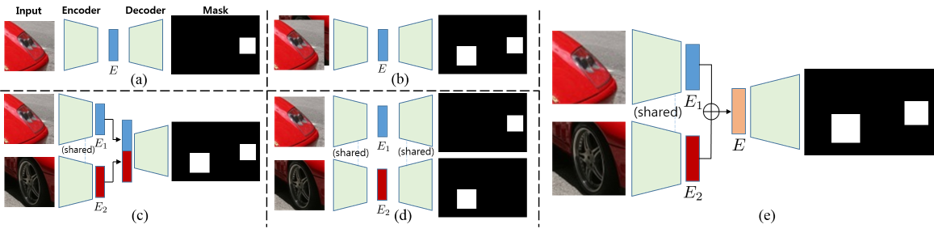}
    \caption{
    Different structures of networks to predict a mask from input patches.
    We choose (e) as our encoder-decoder model.
}
\label{fig:EnDe}
\end{figure*}
Key patches are defined as informative local regions to generate the entire image.
For example, when generating a face image, patches of eyes and a nose
are more informative than those of the forehead and cheeks.
%
Therefore, it would be better for the key patches to contain important parts that can describe objects in a target class.
However, it is not desirable to manually fix the categories of key patches since objects in different classes are composed of different parts.
To address this issue, we use the objectness score from the Edgebox algorithm \cite{zitnick2014edge} to detect key patches.
It can detect key patches of objects in general classes in an unsupervised manner.
In addition, we discard detected patches with extreme sizes or aspect ratios.
Figure \ref{fig:keyPatches} shows examples of detected key patches from various objects and scenes.
Overall, the detected regions from these object classes are fairly informative.
We sort candidate regions by the objectness score and feed the top $N$ patches to the proposed network.
In addition, the training images and corresponding key patches are augmented using a random left-right flip with the equal probability.

\vspace{-2mm}
\subsection{Part Encoding Network}
\vspace{-2mm}
The structure of the generator is based on the encoder-decoder network \cite{hinton2006reducing}.
It uses convolutional layers as an encoder to reduce the dimension of the input data until the bottleneck layer.
Then, transposed convolutional layers upsample the embedded vector to its original size.
For the case with a single input, the network has a simple structure as shown in Figure \ref{fig:EnDe}(a).
For the case with multiple inputs as considered in the proposed network, there are many possible structures.
We examine four cases in this work.

The first network is shown in Figure \ref{fig:EnDe}(b), which uses depth-concatenation of multiple patches.
This is a straightforward extension of the single input case.
However, it is not suitable for the task considered in this work.
Regardless of the order of input patches,
the same mask should be generated when
the patches have the same appearance.
Therefore, the embedded vector $E$
must be the same for all different orderings of inputs.
Nevertheless, the concatenation causes the network to depend on the ordering,
while key patches have an arbitrary order since
they are sorted by the objectness score.
In this case, the part encoding network cannot learn proper filters.
The same issue arises in the model in Figure \ref{fig:EnDe}(c).
On the other hand, there are different issues with the network in Figure \ref{fig:EnDe}(d).
While it can solve the ordering issue, it predicts a mask of each input independently, which
is not desirable as we aim to predict masks jointly.
The network should consider the appearance of both input patches to predict positions.
To address the above issues, we propose to use the network in Figure \ref{fig:EnDe}(e).
It encodes multiple patches based on a Siamese-style network and summarizes all results in a single descriptor by the summation, i.e., $E=E_1+...+E_N$.
Due to the commutative property, we can predict a mask jointly, even if inputs have an arbitrary order.
In addition to the final bottleneck layer, we use all convolutional feature maps in the part encoding network to construct
U-net \cite{ronneberger2015u} style architectures as shown in Figure \ref{fig:proposedNet}.

\vspace{-2mm}
\subsection{Mask Prediction Network}
\vspace{-2mm}
The U-net is an encoder-decoder network that has skip connections
between $i$-th encoding layer and $(L-i)$-th decoding layer, where $L$ is the total number of layers.
%
It directly feeds the information from an encoding layer to its corresponding decoding layer.
Therefore, combining the U-net and a generation network is effective
when the input and output share the same semantic~\cite{isola2016image}.
In this work, the shared semantic of input patches and the output mask is the target image.

We pose the mask prediction as a regression problem.
Based on the embedded part vector $E$, we use transposed convolutional layers with
a fractional stride \cite{radford2015unsupervised} to upsample the data.
The output mask has the same size as the target image and has a value between 0 and 1 at each pixel.
Therefore, we use the sigmoid activation function at the last layer.
The detailed configurations are presented in Table \ref{tab:layers}.

The spatial loss, $\mathcal{L}_S$, is defined as follows:
\begin{equation}\label{eq:spatial_loss}
  \mathcal{L}_{S}(G_\mathcal{M}) = \mathbb{E}_{\mathbf{x}\sim p_{data}(\mathbf{x}),M\sim p_{data}(M)}[\|G_{\mathcal{M}}(\mathbf{x})-M\|_1].
\end{equation}
We note that other types of losses, such as the $l_2$-norm,
or more complicated network structures, such as GAN, have been evaluated
for mask prediction, and similar results are achieved by these alternative options.

\begin{table*}[t]
    \centering
    \footnotesize
    \caption{Details of each network. \# Filter is the number of filters. BN is the batch normalization. Conv denotes a convolutional layer. F-Conv denotes a transposed convolutional layer that uses the fractional-stride.}
    \vspace{-2mm}
\subfloat[Details of the \{part encoding, discriminator\} network]{
    \begin{tabular}{|c|c|c|c|c|c|}
    \hline
    Layer & \# Filter & Filter Size & Stride & Pad & BN \\
    \hline
    Conv. 1 & 64   & $5\times5\times3$   & 2 & 2 & $\times$ \\
    Conv. 2 & 128  & $5\times5\times64$  & 2 & 2 & $\bigcirc$ \\
    Conv. 3 & 256  & $5\times5\times128$ & 2 & 2 & $\bigcirc$ \\
    Conv. 4 & 512  & $5\times5\times256$ & 2 & 2 & $\bigcirc$ \\
    Conv. 5 & 1024 & $5\times5\times512$ & 2 & 2 & $\bigcirc$ \\
    Conv. 6 & \{100,1\} & $1\times1\times1024$ & 1 & 0 & \{$\bigcirc$,$\times$\} \\
    \hline
    \end{tabular}%
  \label{tab:EnDi}%
}
\subfloat[Details of the \{mask prediction, image generation\} network]{
    \begin{tabular}{|c|c|c|c|c|c|}
    \hline
    Layer & \# Filter & Filter Size & Stride & Pad & BN \\
    \hline
    Conv. 1 & $4\times4\times1024$ & $1\times1\times\{100,200\}$ & 1 & 0 & $\bigcirc$ \\
    F-Conv. 2 & 1024 & $5\times5\times512$     & 1/2 & - & $\bigcirc$ \\
    F-Conv. 3 & 512  & $5\times5\times256$     & 1/2 & - & $\bigcirc$ \\
    F-Conv. 4 & 256  & $5\times5\times128$     & 1/2 & - & $\bigcirc$ \\
    F-Conv. 5 & 128  & $5\times5\times64$      & 1/2 & - & $\bigcirc$ \\
    F-Conv. 6 & 64   & $5\times5\times\{1,3\}$ & 1/2 & - & $\times$ \\
    \hline
    \end{tabular}%
  \label{tab:DeGe}%
}
\label{tab:layers}
\end{table*}

\begin{figure}[!t]
    \centering
    \includegraphics[width=0.9\linewidth]{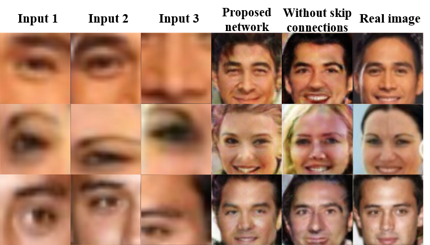}
    \caption{
    Sample image generation results on the CelebA dataset.
    Images are generated based on the network in Figure \ref{fig:proposedNet}.
    Generated images are sharper and realistic with the skip connections.
}
\label{fig:with_or_without_U}
\end{figure}
\vspace{-2mm}
\subsection{Image Generation Network}
\vspace{-2mm}
We propose a double U-net structure for the image generation task as shown in Figure \ref{fig:proposedNet}.
It has skip connections from both the part encoding network and mask generation network.
In this way, the image generation network can communicate with other networks.
This is critical since the generated image should consider the appearance and locations of input patches.
Figure \ref{fig:with_or_without_U} shows generated images with and without the skip connections.
It shows that the proposed network improves the quality of generated images.
In addition, it helps to preserve the appearances of input patches.

Based on the generated image and predicted mask, we define the appearance loss $\mathcal{L}_A$ as follows:
\begin{equation}\label{eq:appearance_loss}
\begin{aligned}
  \mathcal{L}_{A}(G_\mathcal{M},G_\mathcal{I})
  = {} & \mathbb{E}_{\mathbf{x},y\sim p_{data}(\mathbf{x},y), z\sim p_z(z), M\sim p_{data}(M)} \\
       & [\|G_\mathcal{I}(\mathbf{x},z)\otimes G_\mathcal{M}(\mathbf{x}) - y\otimes M\|_1],
\end{aligned}
\end{equation}
where $\otimes$ is an element-wise product.

\vspace{-2mm}
\subsection{Real-Fake Discriminator Network}
\vspace{-2mm}
A simple discriminator can be trained to distinguish real images from fake images.
However, it has been shown that
a naive discriminator may cause artifacts \cite{shrivastava2016learning} or
network collapses during training \cite{metz2016unrolled}.
To address this issue, we propose a new objective function as follows:
\begin{equation}\label{eq:realfake_loss}
\begin{aligned}
  \mathcal{L}_{R}(G_\mathcal{I}, D&) = {} \mathbb{E}_{y\sim p_{data}(y)}[\log D(y)] + \\
  & \mathbb{E}_{\mathbf{x},y,y'\sim p_{data}(\mathbf{x},y,y'), z\sim p_z(z), M\sim p_{data}(M)} \\
  & [\log(1-D(G_\mathcal{I}(\mathbf{x},z))) + \\
  & \log(1-D(M\otimes G_\mathcal{I}(\mathbf{x},z) + (1-M)\otimes y)) + \\
  & \log(1-D((1-M)\otimes G_\mathcal{I}(\mathbf{x},z) + M\otimes y)) + \\
  & \log(1-D(M\otimes y' + (1-M)\otimes y)) + \\
  & \log(1-D((1-M)\otimes y' + M\otimes y))],
\end{aligned}
\end{equation}
where $y'$ is a real image randomly selected from the outside of the current mini-batch.
When the real image $y$ is combined with the generated image $G_\mathcal{I}(\mathbf{x},z)$ (line 4-5 in (\ref{eq:realfake_loss})),
it should be treated as a fake image as it partially contains the fake image.
%
When two different real images $y$ and $y'$ are combined (line 6-7 in (\ref{eq:realfake_loss})),
it is also a fake image although both images are real.
It not only enriches training data but also strengthens discriminator by feeding difficult examples.

\begin{figure*}[!t]
    \centering
    \subfloat[CelebA dataset]{\includegraphics[width=0.48\linewidth]{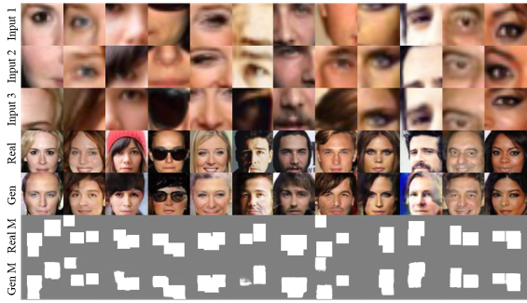}%
    \label{fig:genCeleb}}
    \quad
    \subfloat[Waterfall dataset]{\includegraphics[width=0.48\linewidth]{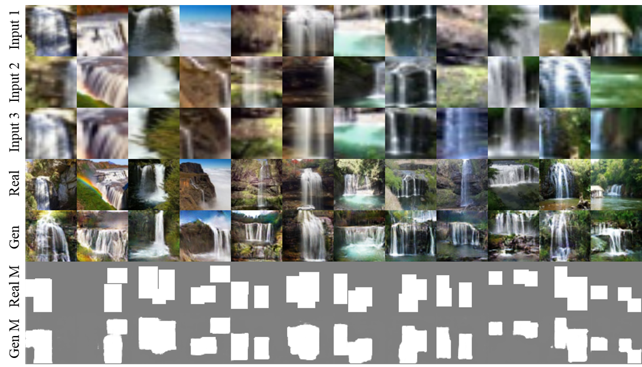}%
    \label{fig:genWaterfall}}
    \hfil
    \vspace{-2mm}
    \subfloat[CompCars dataset]{\includegraphics[width=0.48\linewidth]{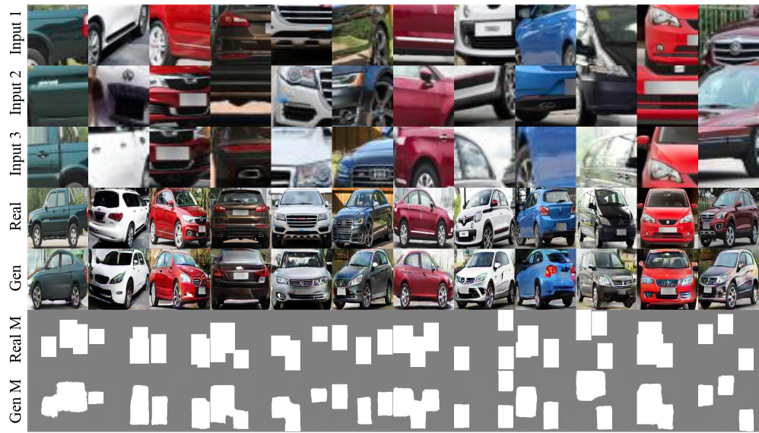}%
    \label{fig:genCompCars}}
    \quad
    \subfloat[Stanford Cars dataset]{\includegraphics[width=0.48\linewidth]{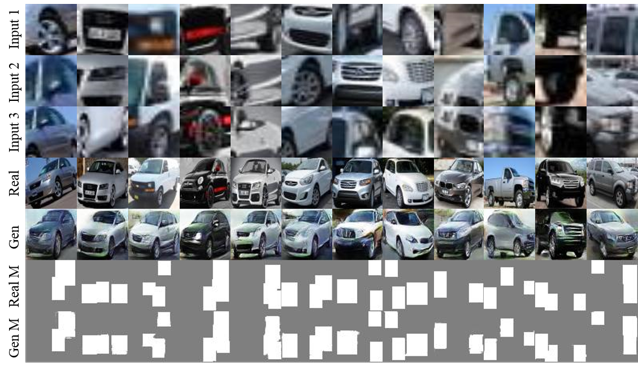}%
    \label{fig:genStanford}}
    \hfil
    \vspace{-2mm}
    \subfloat[Flower dataset]{\includegraphics[width=0.48\linewidth]{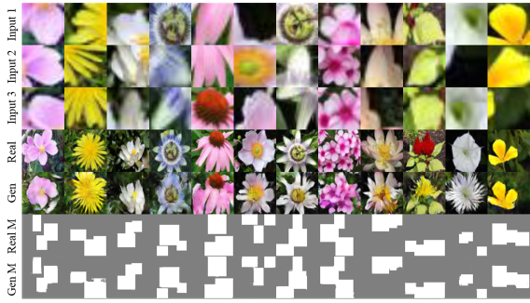}%
    \label{fig:genFlower}}
    \quad
    \subfloat[Ceramic dataset]{\includegraphics[width=0.48\linewidth]{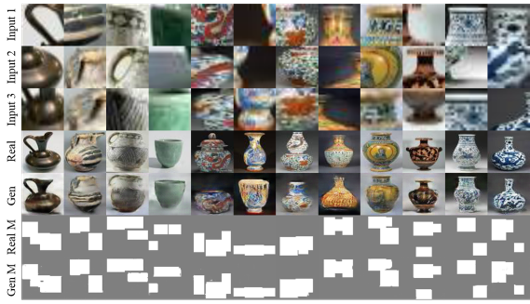}%
    \label{fig:genCeramic}}
    \hfil
    \caption{
    Examples of generated masks and images on six datasets.
    The generated images for each class are shown in 12 columns.
    Three key local patches (Input 1, Input 2, and Input 3) from a real image (Real).
    The key parts are top-3 regions in terms of the objectness score.
    Given inputs, images (Gen) and masks (Gen M) are generated.
    Real M is the ground truth mask.
    }
    \label{fig:gen}
\end{figure*}

\vspace{-2mm}
\section{Experiments}
\vspace{-2mm}

For all experiments, images are resized to the minimum length of 128 pixels on the width or height.
All key part candidates are obtained using the Edgebox algorithm \cite{zitnick2014edge}.
We reject candidate boxes that are larger than 25\% or smaller than 5\% of the image size unless otherwise stated.
After that, the non-maximum suppression is applied to remove candidates that are too close with each other.
Finally, the image and top $N$ candidates are resized to the target size, $128\times128\times3$ pixels for the CompCars dataset or $64\times64\times3$ pixels for other datasets, and fed to the network.
The $\lambda_1$ and $\lambda_2$ are decreased from $10^{-2}$ to $10^{-4}$ as the epoch increases.

Table \ref{tab:layers} shows detailed description of the proposed network for $128\times128\times3$ pixels image.
The input parts are encoded into a 100-dimensional vector $E$.
A mask is predicted using $E$,
while an image is generated based on a 200-dimensional vector which is a concatenation of $E$ and a 100-dimensional random noise vector $z$.
The part encoding network uses the leaky ReLU \cite{maas2013rectifier} with a slope of 0.2 as an activation function.
The discriminator uses the same leaky ReLU except for the last layer which uses a sigmoid function.
The mask prediction and image generation networks use ReLU except for the last layer which uses a sigmoid function and $\tanh$, respectively.
The filters in the network are initialized with a zero mean Gaussian distribution with a standard deviation of 0.02.

We train the network with a learning rate of 0.0002.
As the epoch increases, we decrease $\lambda1$ and $\lambda2$ in (\ref{eq:loss}).
With this training strategy, the network focuses on predicting a mask in the beginning,
while it becomes more important to generate realistic images in the end.
The mini-batch size is 64, and the momentum of the Adam optimizer \cite{kingma2014adam} is set to 0.5.
During training, we first update the discriminator network and then update the generator network twice.
More results are available in the supplementary material.
All the source code and datasets will be made available to the public.
%

\subsection{Datasets}
\vspace{-2mm}
The CelebA dataset \cite{liu2015faceattributes} contains
202,599 celebrity images with large pose variations and background clutters (see Figure~\ref{fig:gen}(a)).
There are 10,177 identities with various attributes, such as
eyeglasses, hat, mustache, and facial expressions.
We use aligned and cropped face images of $108\times108$ pixels.
The network is trained for 25 epochs.

The flower dataset \cite{nilsback2008automated} consists of 102 flower categories (see Figure~\ref{fig:gen}(e)).
There is a total of 8,189 images, and each class has between 40 and 258 images.
The images contain large variations in the scale, pose, and lighting condition.
We train the network for 800 epochs.

There are two car datasets \cite{yang2015large,krause20133d} used in this paper.
The CompCars dataset \cite{yang2015large} includes images from two scenarios: the web-nature and surveillance-nature (see Figure~\ref{fig:gen}(c)).
The web-nature data contains 136,726 images of 1,716 car models, and the surveillance-nature data contains 50,000 images.
The network is trained for 50 epochs.
The Stanford Cars dataset \cite{krause20133d} contains 16,185 images of 196 classes of cars (see Figure~\ref{fig:gen}(d)).
They have different lighting conditions and camera angles.
Furthermore, a wide range of colors and shapes, e.g., sedans, SUVs, convertibles, trucks,
are included.
The network is trained for 400 epochs.

The waterfall dataset consists of 15,323 images taken from various viewpoints (see Figure~\ref{fig:gen}(b)).
It has different types of waterfalls as images are collected from the internet.
It also includes other objects such as trees, rocks, sky, and ground,
as images are obtained from natural scenes.
For this dataset, we allow tall candidate boxes, in which the maximum height is 70\% of the image height, to catch long water streams.
The network is trained for 100 epochs.

The ceramic dataset is made up of 9,311 side-view images (see Figure~\ref{fig:gen}(f)).
Images of both Eastern-style and Western-style potteries are collected from the internet.
The network is trained for 800 epochs.

\begin{figure}[!t]
    \centering
    \includegraphics[width=1.0\linewidth]{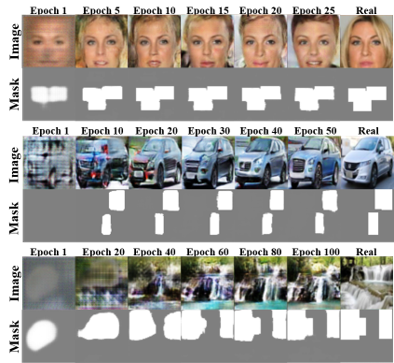}
    \caption{
    Sample generated masks and images at different epochs.
}
\label{fig:epochGen}
\end{figure}

\vspace{-2mm}
\subsection{Image Generation Results}
\vspace{-2mm}
Figure \ref{fig:epochGen} shows generation results as the training epoch is increased.
At the start, the network focuses on predicting a good mask.
As the epoch is increased, input parts become sharper.
At the end of the epoch, the network concentrates on generating realistic images.
In the case of the CelebA dataset, it is relatively easy to find the mask since the images of this dataset are aligned.
On the other hand, for other datasets, it takes more epochs to find a good mask.
The results show that the masked regions have similar appearances while other regions are changed in a way to make realistic holistic images.

Figure \ref{fig:gen} shows image generation results of different object classes.
Each input has three key patches from a real image and we show both generated and original ones for visual comparisons.
For all datasets, which contain challenging objects and scenes,
the proposed algorithm is able to generate realistic images.
The subject of the generated face images using the CelebA dataset
in Figure~\ref{fig:gen}(a) may have different gender (column 1 and 2),
wear a new beanie or sunglasses (column 3 and 4), and
become older, chubby, and with new hairstyles (column 5-8).
Even when the input key patches are concentrated on the left or right sides,
the proposed algorithm can generate realistic images (column 9 and 10).
In the CompCars dataset, the shape of car images is mainly generated based on the
direction of tire wheels, head lights, and windows.
For some cases, such as column 2 in Figure~\ref{fig:gen}(c),
input patches can be from both left or right directions and the generation results can be flipped.
It demonstrates that the proposed algorithm is flexible
since the correspondence between the generated mask and input patches,
e.g., the left part of the mask corresponds to the left wheel patch, is not needed.
Due to the small number of training samples compared to the CompCars dataset,
the results of the Stanford Cars dataset are less sharp but still realistic.
For the waterfall dataset, the network learns how to draw
a new water stream (column 1),
a spray from the waterfall (column 3),
or other objects such as rock, grass, and puddles (column 10).
In addition, the proposed algorithm can help restoring broken pieces of ceramics found in ancient ruins (see Figure~\ref{fig:gen}(f)).

\begin{figure}[t]
    \centering
    \includegraphics[width=1.0\linewidth]{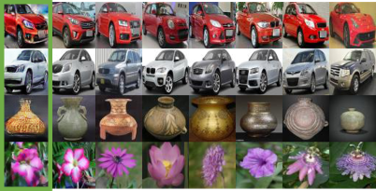}
    \caption{
    For each generated image in the green box, nearest neighbors in the corresponding training dataset are displayed.
}
\label{fig:near}
\end{figure}

Figure \ref{fig:near} shows nearest neighbors of generated images.
We measure the Euclidean distance between the generated image and images in the training set to define neighbors.
The generated images are visually similar to real images in the training set, but have clear differences.

\begin{figure}[t]
    \centering
    \includegraphics[width=1.0\linewidth]{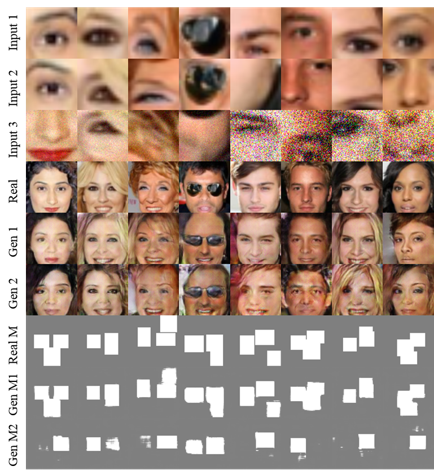}
    \caption{
    Examples of generated results when the input image contains noises.
    We add a Gaussian noise at each pixel of Input 3.
    Gen 1 and Gen M1 are generated without noises.
    Gen 2 and Gen M2 are generated with noises.
}
\label{fig:noise}
\end{figure}
\begin{figure*}[!t]
    \centering
    \includegraphics[width=1.0\linewidth]{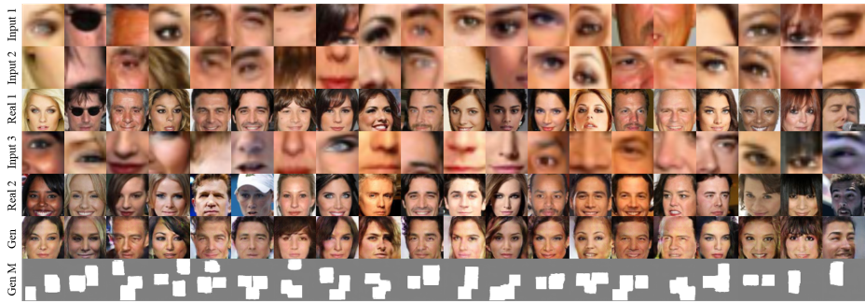}
    \caption{
    Results of the proposed algorithm on the CelebA dataset when input patches are came from other images.
    Input 1 and Input 2 are patches from Real 1. Input 3 is a local region of Real 2.
    Given inputs, the proposed algorithm generates the image (Gen) and mask (Gen M).
}
\label{fig:combi}
\end{figure*}

Figure \ref{fig:noise} shows the results when input patches are degraded by noises.
We apply the mean zero Gaussian noise at each pixel of the third input patch with
the standard deviation of 0.1 (column 1-4) and 0.5 (column 5-8).
The results show that the proposed algorithm is able to deal with certain amount of noise when generating realistic images.

Figure \ref{fig:combi} shows generated images and masks when input patches are obtained from different persons.
The results show that the proposed algorithm can handle a wide scope of input patch variations.
%
For example, inputs contain different skin colors in the first column.
In this case, it is not desirable to exactly preserve inputs since
it will generate a face image with two different skin colors.
The proposed algorithm generates an image with a reasonable skin color as well as the overall shape.
Other cases include with or without sunglasses (column 2), different skin textures (column 3), hairstyle variations (column 4 and 5), and various expressions and orientations.
Despite large variations, the proposed algorithm is able to generate realistic images.

\begin{figure}[t]
    \centering
    \includegraphics[width=1.0\linewidth]{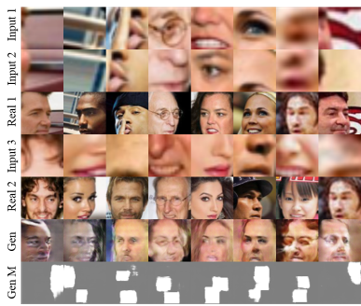}
    \caption{
    Examples of failure cases of the proposed algorithm.
}
\label{fig:fail}
\end{figure}

Figure \ref{fig:fail} shows failure cases of the proposed algorithm.
It is difficult to generate images when detected key input patches include less informative regions (column 1 and 2) or rare cases (column 3).
In addition, when input patches have conflicting information, e.g., the same nose-mouth patches that have different orientations,
the proposed algorithm is not able to generate realistic images (column 4, 5, and 6).
Furthermore, it becomes complicated when the inputs are low-quality patches (column 7 and 8).
%
We note these issues can be alleviated with additional pre-processing modules.

\vspace{-2mm}
\section{Conclusions}
\vspace{-2mm}
We introduce a new problem of generating images based on local patches without geometric priors.
Local patches are obtained using the objectness score to retain informative parts of the target image in an unsupervised manner.
We propose a generative network to render realistic images from local patches.
The part encoding network embeds multiple input patches using a Siamese-style convolutional neural network.
Transposed convolutional layers with skip connections from the encoding network are used to predict a mask and generate an image.
The discriminator network aims to classify the generated image and the real image.
The whole network is trained using the spatial, appearance, and adversarial losses.
Extensive experiments show that the proposed network can generate realistic images of challenging objects and scenes.
As humans can visualize a whole scene with a few visual cues, the proposed network can generate realistic images based on given unordered image patches.

{\small
\bibliographystyle{ieee}
\bibliography{arxiv2017_gan}
}

\clearpage
\onecolumn
\section{Supplementary Material}
\subsection{Image Generation from Parts of Different Cars}
\label{sec:combi}

\begin{figure*}[t]
    \centering
    \includegraphics[width=0.88\linewidth]{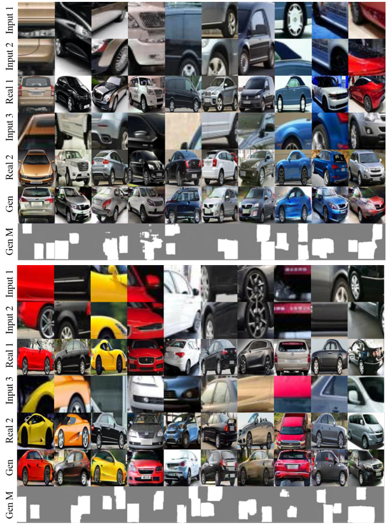}
    \caption{
    Results of the proposed algorithm on the CompCars dataset when input patches are from different cars.
    Input 1 and Input 2 are patches from Real 1.
    Input 3 is a local region of Real 2.
    Given inputs, the proposed algorithm generates the image (Gen) and mask (Gen M).
    The size of the generated image is of $128\times128$ pixels.
}
\label{fig:combicar}
\end{figure*}

Figure \ref{fig:combicar} shows generated images and masks when input patches are from different cars.
Overall, the proposed algorithm generates reasonable images despite large variations of input patches.

\subsection{Image Generation from a Different Number of Patches}
\label{sec:input2}

In the manuscript, we show image generation with three local patches using the proposed algorithm.
Figure \ref{fig:input2} shows generated images based on two local patches.
The results show that the network can be trained with different number of input patches.

%

\begin{figure*}[!t]
    \centering
    \subfloat[]{\includegraphics[width=1.0\linewidth]{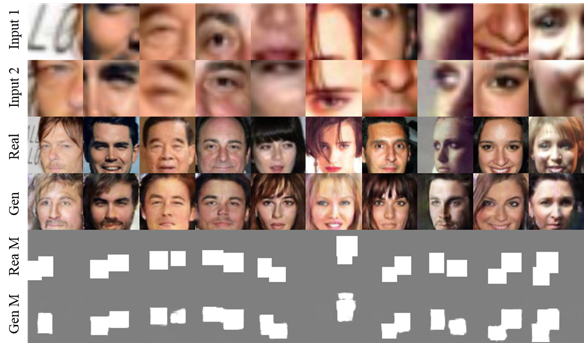}
    \label{fig:input2_1}}
    \hfil
    \subfloat[]{\includegraphics[width=1.0\linewidth]{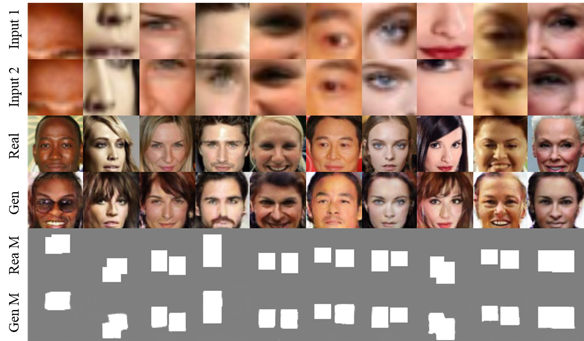}\
    \label{fig:input2_2}}
    \hfil
    \phantomcaption
\end{figure*}

\begin{figure*}[!t]
    \ContinuedFloat
    \subfloat[]{\includegraphics[width=1.0\linewidth]{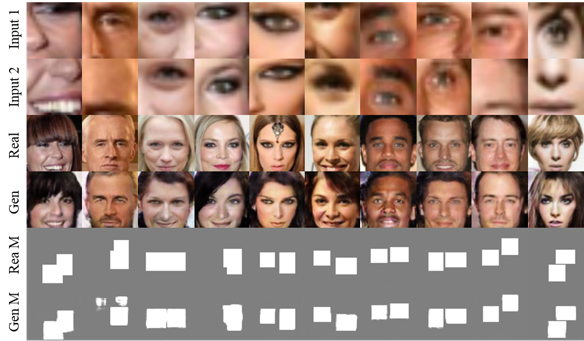}\
    \label{fig:input2_3}}
    \caption{
    Image generation results with two input patches.
    Input 1 and 2 are local patches from the image Real.
    }
    \label{fig:input2}
\end{figure*}

\subsection{Image Generation using an Alternative Objective Function}
\label{sec:self}

In order to demonstrate the effectiveness of (4) in the paper,
we show generation results in Figure \ref{fig:basicD} using the following objective function:
\begin{equation}\label{eq:basicD}
  \mathcal{L}_{R}(G_\mathcal{I}, D) = {} \mathbb{E}_{y\sim p_{data}(y)}[\log D(y)] +
   \mathbb{E}_{\mathbf{x},y,y'\sim p_{data}(\mathbf{x},y,y'), z\sim p_z(z), M\sim p_{data}(M)}
   [\log(1-D(G_\mathcal{I}(\mathbf{x},z)))].
\end{equation}
Both results are obtained after 25 epochs.
The results show that generated images with (\ref{eq:basicD}) are less realistic compared to the results of (4) in the paper.

\begin{figure*}[t]
    \centering
    \subfloat[]{\includegraphics[width=1.0\linewidth]{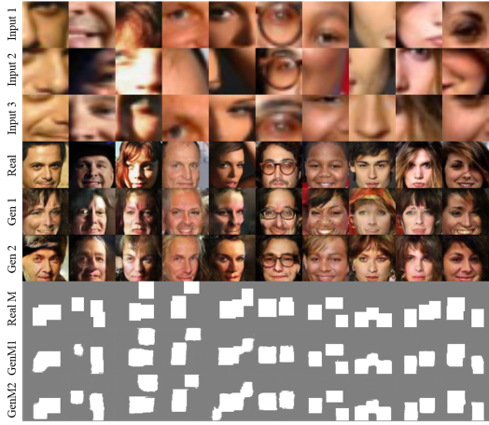}
    \label{fig:basicD2_1}}
\phantomcaption
\end{figure*}

\begin{figure*}[t]
    \ContinuedFloat
    \centering
    \subfloat[]{\includegraphics[width=1.0\linewidth]{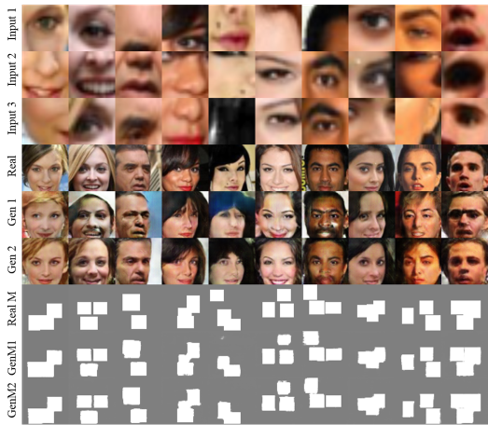}
    \label{fig:basicD2_2}}
    \caption{
    Image generation results on the CelebA dataset.
    Gen 1 and GenM1 are generated by (\ref{eq:basicD}).
    Gen 2 and GenM2 are obtained using (4) in the paper.
    \label{fig:basicD}
}
\end{figure*}

\end{document}